\documentclass{article}

\usepackage{arxiv}

\usepackage[utf8]{inputenc} 
\usepackage[T1]{fontenc}    
\usepackage{hyperref}       
\usepackage{url}            
\usepackage{booktabs}       
\usepackage{amsfonts}       
\usepackage{nicefrac}       
\usepackage{microtype}      
\usepackage{lipsum}

\usepackage{latexsym}
\usepackage{amssymb}
\usepackage[T1]{fontenc}
\usepackage{amsmath}
\usepackage{graphicx}
\usepackage{amsmath}
\usepackage{xcolor,float}
\usepackage{soul}
\usepackage{setspace}
\usepackage{natbib}
\title{WheatNet: A Lightweight Convolutional Neural Network for High-throughput Image-based Wheat Head Detection and Counting}

\author{
  Saeed Khaki \\
  Industrial Engineering Department\\
  Iowa State University\\
  Ames, Iowa 50011, USA\\
  
  \texttt{skhaki@iastate.edu} \\
  
  \And
  
  Nima Safaei\\
  Tippie College of Business\\ University of Iowa\\
  Iowa City, Iowa 52242, USA\\
   \And
 Hieu Pham\\
 Industrial Engineering Department\\
  Iowa State University\\
  Ames, Iowa 50011, USA\\

\And
Lizhi Wang \\
  Industrial Engineering Department\\
  Iowa State University\\
  Ames, Iowa 50011, USA\\
}

\begin{document}
\maketitle

\begin{abstract}
For a globally recognized planting breeding organization, manually-recorded field observation data is crucial for plant breeding decision making. However, certain phenotypic traits such as plant color, height, kernel counts, etc. can only be collected during a specific time-window of a crop's growth cycle. Due to labor-intensive requirements, only a small subset of possible field observations are recorded each season.  To help mitigate this data collection bottleneck in wheat breeding, we propose a novel deep learning framework to accurately and efficiently count wheat heads to aid in the gathering of real-time data for decision making. We call our model WheatNet and show that our approach is robust and accurate for a wide range of environmental conditions of the wheat field. WheatNet uses a truncated MobileNetV2 as a lightweight backbone feature extractor which merges feature maps with different scales to counter image scale variations. Then, extracted multi-scale features go to two parallel sub-networks for simultaneous density-based counting and localization tasks. Our proposed method achieves an MAE and RMSE of 3.85 and 5.19 in our wheat head counting task, respectively, while having significantly fewer parameters when compared to other state-of-the-art methods. Our experiments and comparisons with other state-of-the-art methods demonstrate the superiority and effectiveness of our proposed method. 
\end{abstract}

\keywords{ Wheat head detection; Convolutional neural network; Deep learning; Agriculture; Yield estimation}

\section{Introduction}
Wheat (\textit{Triticum}) is one of the most cultivated crops in the world. In 2019, global wheat production achieved 762 million tonnes, most of which was used for human consumption \citep{reynolds2006applying}. Due to the high nutritional components contained in wheat such as protein, vitamins, fiber, and phytochemicals, the importance of this crop cannot be undersold. The continued development of global wheat is crucial for long-term food security. However, since the 1990s, the rate of increase in wheat yield has slowed \citep{brisson2010wheat,schauberger2018yield}. This is due, in part, to the manual labor requirements needed for traditional wheat breeding. Understanding phenotypic traits for any crop (height, root color, stalk color, kernel counts, etc.) is vital in improving its growth rate and, ultimately, yield. Specifically for wheat, knowing the number of heads per variety is key in estimating yield performance. Combining this information with environmental and genetic factors, plant breeders are able to selectively choose which parents (out of potentially tens of thousands of candidates) to cross to create the next generation of superior wheat varieties \citep{madec2019ear}. However, due to being reliant on intensive, physical, manual labor to count wheat heads human-error is inevitable with studies showing a measurement error of approximately 10\% for this task \citep{hasan2018detection}. Coupled with the ability to only record head count during a specific time-window of a wheat variety's growth cycle makes this task too difficult to complete in a timely manner. In practice, due to limited time and resource constraints,  head count data is collected for only a small subset of all possible wheat varieties - even for a commercial breeding organization.

With the advent of modern large-scale image capturing devices such as unmanned aerial vehicles (UAVs), satellites, and mobile cameras, plant breeders and farmers are now able to capture images of crop fields at a faster rate and store images for later analysis \citep{patel2016agriculture,puri2017agriculture,nguyen2020monitoring, peng2020evaluation, khaki2020yieldnet,khaki2019crop,khaki2020predicting,khaki2019classification}. The adoption of new image capturing technology has the ability to reduce both intensive field labor and time-window restrictions. However, this new process welcomes a different challenge. For any breeding organization or large-scale farmer, who manages thousands of acres of wheat, how can image analysis be performed at scale accurately and efficiently?

To achieve true high throughput plant phenotyping, accurate and automated techniques must be adopted. With the aid of increased GPU computing power, automating image analysis tasks is now possible and already being used in the agricultural sector for image classification and object detection and counting. For example, well establish deep learning architectures  such as AlexNet \citep{krizhevsky2012imagenet}, LeNet \citep{lecun2015deep}, and VGG-16 \citep{simonyan2014deep} have been applied to classify diseases in both fruit and vegetables \citep{mohanty2016using,wang2017automatic}. Moreover, ResNet-50 has been utilized extensively for object detection in count leaves and sorghum heads\citep{giuffrida2018pheno, mosley2020image}. Custom deep learning models have also been proposed to detect corn tassels by combining convolutional neural networks and local counts regression \citep{lu2017tasselnet}. Recently numerous approaches have been proposed for image-based corn kernel counting and image-based plant stand counting \citep{khaki2019cnn,khaki2020high, khaki2020convolutional,  khaki2021deepcorn}. These approaches follow a similar framework to crowd counting models due to the density of objects in the images. Aside from research into model building for image analysis in agriculture, publically available, annotated, crop datasets have been made available to researchers so that they may contribute to the growing area of computer vision and agriculture. In addition to deep learning techniques, recent literature has been favorable to the intersection of general machine learning and image processing techniques and agriculture \citep{singh2016machine,naik2017real, dobbels2019soybean, pothen2020detection}. With the rapid combination of deep learning and plant breeding, there is hope that the data collection bottleneck can be mitigated to help maximize the growth potential of crops fueling our world. For the curious reader, we direct them to a thorough review of image-based plant phenotyping by \cite{jiang2020convolutional}.

While recent work is favorable to deep learning approaches in agriculture, research combining image-based analytics and wheat is relatively untouched. To engage future researchers to help tackle the problem of wheat head detection, \cite{david2020global} released a large, annotated, public dataset that allows researchers to study this problem. Making use of this open-source dataset, \cite{gong2021real} applied an object detection framework called YOLOv4 to localize and classify wheat heads using bounding boxes. Although their work has shown success their approach did not consider a complete end-to-end framework to allow a scalable, deployable technique to be widely adopted by breeders, agronomists, and farmers. Indeed, to obtain high throughput phenotyping for wheat head detection and counting, a mechanism must be created that is easily deployable in-field and with minimal inference time. It is the goal of this paper to make use of this available wheat head dataset to enable true high throughput phenotyping to aid in large-scale data capturing to ultimately improve decision making of either a plant breeder or a large-scale farmer.

One approach to increase wheat yield is to provide agronomists, breeders, and farmers with real-time, precise data to estimate yield during the growing season. This mechanism would enable farmers to make real-time management decisions to minimize their loss while maximizing yield. Being able to estimate in-field yield quickly enables farmers to be proactive in applying optimal management practices (when to apply fertilizer, fungicide, nitrogen, etc.) to maximize growth potential \citep{meier2020management, rhebergen2020closing, shahhosseini2020forecasting, imran2021carbon}. Currently, to estimate wheat yield, farmers must manually count the number of wheat heads in a given unit area and then input that count into a well-established agronomic formula to estimate wheat yield \citep{lupton1987history, nebwheat}. However, with the existence of  150-200 wheat heads in an area coupled with the need to repeat the process for different parts of the field, this manual task becomes challenging to complete in a timely manner. Conversely for a global, commercial, breeding organization having a deployable tool to automate in-field wheat head counting allows plant breeders to acquire an important phenotypic trait for a large number of wheat varieties with reduced labor costs and without error in manual counting from humans. Because important, industry-defining decisions are made utilizing this information, it is pivotal that phenotypic traits are recorded accurately and consistently. Plant breeders use wheat head count in their decision making process to determine which wheat varieties should be crossed to generate a new, superior offspring, and, in the end, sold to farmers. Regardless of the end-user, if a deployable, efficient, and effective framework existed to allow for true high throughput phenotyping of wheat heads, additional vital information can be gathered to help maximize growth potential and, ultimately, crop yield.

Given the need for a system to accurately and efficiently count wheat heads, we propose a novel approach using a deep learning framework we call WheatNet. Our novel methodology takes a combination of object localization and crowd counting techniques. Specifically, we propose a novel framework with the following properties:

\begin{enumerate}
    \item A robust wheat head counting model that makes no assumption on the image scale, orientation, size, lighting conditions, or environmental conditions.
    \item The use of a lightweight encoder and decoder that significantly reduces the number of learned parameters allowing our proposed method to easily be deployed on mobile devices for in-field usage.
    \item Adopts two parallel sub-networks for simultaneous density-based counting and localization which mutually strengthen each other for improving prediction accuracy.
    \item Fast, real-time inference for fast yield estimation and high-throughput phenotyping.
    \item The use of point-level annotations for both counting and localization that is considerably less labor-intensive and time-consuming compared to box-level annotations.
\end{enumerate}

\section{Methodology}\label{method}

This study aims to count and detect wheat heads in an image of a wheat field captured between 1.8 m and 3 m above the ground to use for wheat yield estimation. Our proposed method, WheatNet, uses only point-level annotations to do both counting and localization tasks where the center of the heads are labeled with points. Specifically, our proposed method uses point-level annotations to estimate both the center point and the number of wheat heads. Compared to box-level annotations, annotating an image with points is less labor-intensive and time-consuming. Moreover, given the density of objects, the use of point-level annotations is more practical especially in domains such as agriculture where publicly available annotated datasets are scarce.

\begin{figure}[H]
    \centering
    \includegraphics[scale=0.10]{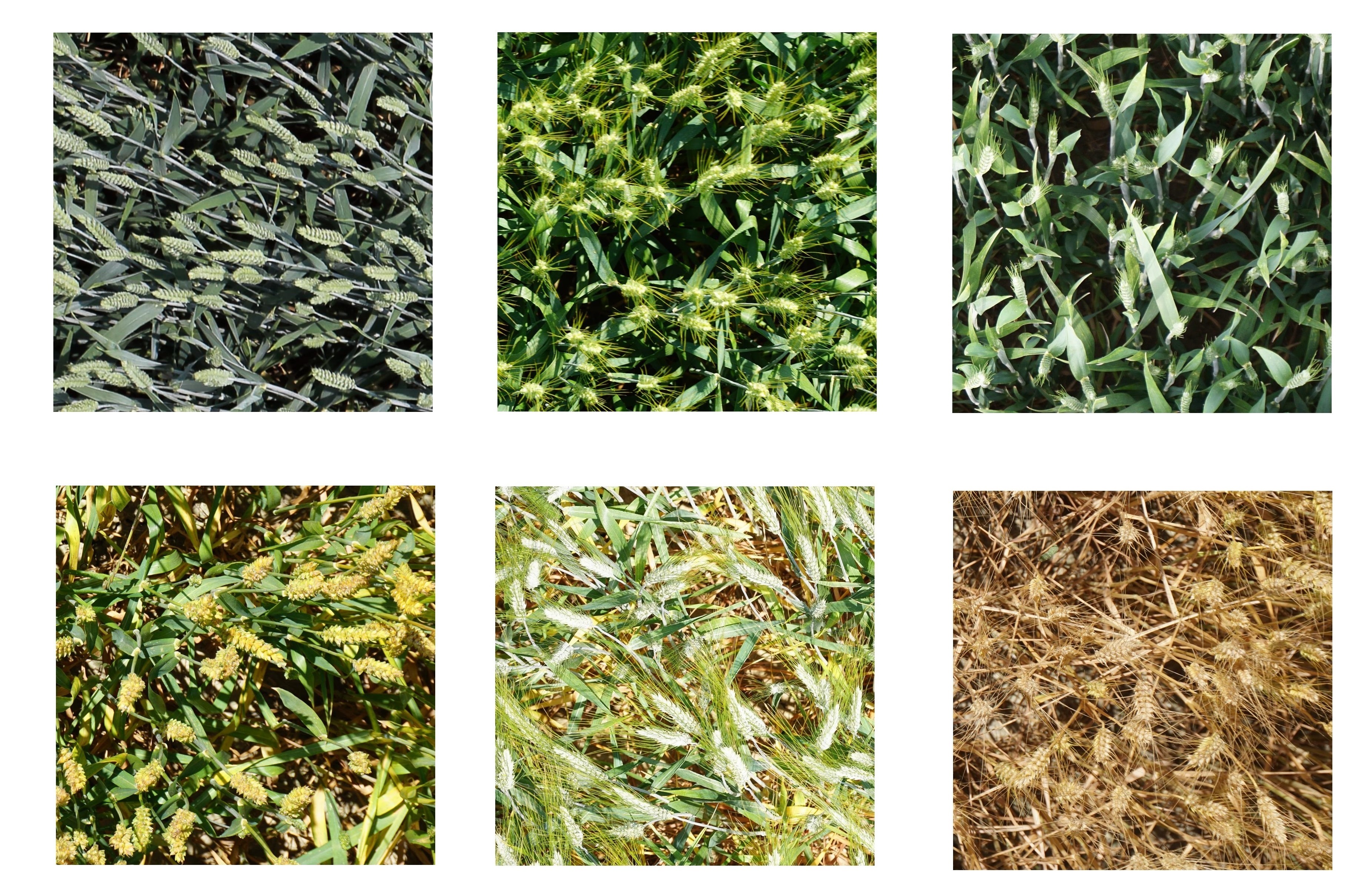}
    \caption{The image illustrates high variability in the wheat genotypes, head orientations, development stages, and image scale.}
    \label{fig:wheat_heads}
\end{figure}

Accurate wheat head localization and counting are challenging due to multiple factors: (1) large variations in wheat heads' shapes and color due to genotype differences, (2) high variability in observational conditions, head orientation, image scales, lighting conditions, and development stages, and (3) overlaps between wheat heads. Figure \ref{fig:wheat_heads} displays different wheat heads which illustrate these challenging factors. Our proposed method is inspired by crowd counting methods which estimate the number of (typically dense) objects in an image such as people \citep{liu2019high,liu2019recurrent,gao2020cnn}. Similar to dense groups of people, our proposed method localizes the wheat heads by estimating their center points.


\subsection{Network Architecture}

\begin{figure}[H]
    \centering
    \includegraphics[scale=0.50]{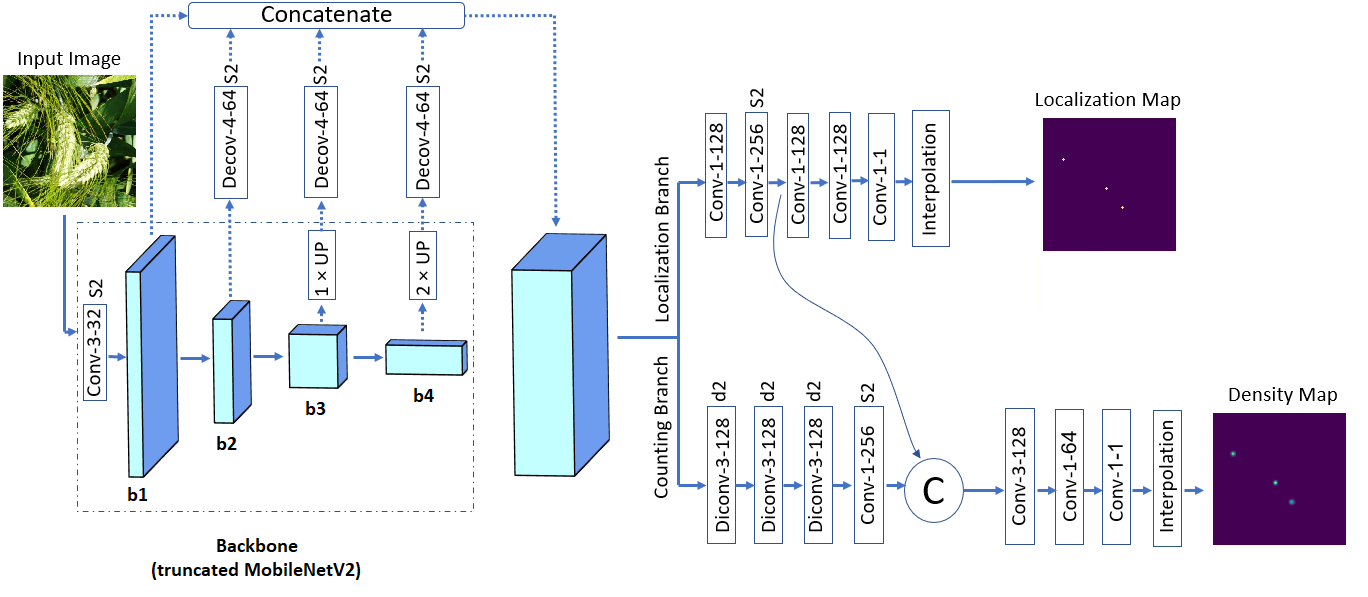}
    \caption{The network architecture of WheatNet. The parameters of the convolutional layers are denoted as “(convolution type)-(kernel size)-(number of filters)”. Conv, Decon, and Diconv stand for standard convolution, deconvolution, and dilated convolution, respectively. “Up” denotes nearest upsampling layer. The amount of stride for all layers is 1 except for layers with “S2” notation for which we use a stride of 2. The dilation rate is denoted as “d”, which is set to be 2 in our model. The padding type is “same” for all layers. \textcircled{\raisebox{-0.8pt}{c}} denotes matrix concatenation. “b” refers to the bottleneck blocks of MobileNetV2. ReLU activation function is used in all layers except the last layers of the network. Here, we zoomed the input image for clear visualization.}
    \label{fig:network_arch}
\end{figure}

For accurately estimating the number of wheat heads and their corresponding centers, our proposed method has two branches for density-based counting and localization, respectively. Our proposed method uses a truncated MobileNetV2 \citep{sandler2018mobilenetv2} as an encoder for feature extraction. Figure \ref{fig:network_arch} outlines the network architecture of WheatNet. MobileNetV2 is a mobile architecture with a substantially reduced memory footprint and improved classification accuracy. MobileNetV2 is based on an inverted residual structure with linear bottlenecks where bottlenecks are mainly composed of depthwise separable convolutions \citep{sandler2018mobilenetv2}. Depthwise separable convolutions are essential parts of efficient mobile neural network architectures that factorize a full convolutional operation to depthwise and pointwise convolution operations \citep{sandler2018mobilenetv2}.

Our proposed method uses a truncated MobileNetV2 as its backbone feature extractor to significantly reduce the model parameters to make it well-suited for mobile applications and deployment. We use the first four bottlenecks of MobileNetV2 and remove the last three bottlenecks as we find that it improves the prediction accuracy while decreasing the computation time. Table \ref{tab:backbone_MB} shows the detailed structure of the truncated MobileNetV2 model. To counter the scale variations and perspective change in images and extract features that include the context information of multiple scales, we merge feature maps from the output of four bottlenecks of MobileNetV2 to obtain multi-scale feature maps. To be able to concatenate feature maps received from MobileNetV2 bottlenecks, we increase the spatial dimensions of feature maps to have the same size as the largest feature map (first feature map) because the spatial dimensions of feature maps are different. As a result, we increase the spatial dimensions of feature maps to have the same size as the largest feature map. We use a deconvolution layer and nearest upsampling layer to increase the spatial dimensions before concatenation. Then, the extracted multi-scale features go to the counting branch and localization branch for density-based counting and center point localization, respectively.

\begin{table}[H]
    \centering
    \begin{tabular}{c c c c c c}
    \hline
      Operator  & Output size & $t$ &$c$ & $n$& $s$\\
      \hline
      image   & 1 & -& -&-\\
      
      conv2d &1/2&-&32&1&2\\
      
      bottleneck &1/2 & 1&48&1&1\\
      
      bottleneck &1/4 & 6&64&2&2\\
      
      bottleneck &1/8 & 6&160&3&2\\
      
      bottleneck &1/16 & 6&256&4&2\\
      \hline

    \end{tabular}
    \caption{Truncated MobileNetV2 backbone architecture. $t$, $c$, $n$, and $s$ denote the expansion factor, number of output channels, repeating factor, and stride, respectively.}
    \label{tab:backbone_MB}
\end{table}

\textbf{Counting branch:}

The counting branch estimates the density map given an image of wheat heads where the integral over any region in the density map gives the number of wheat heads within that region. Motivated by CSRNet \citep{li2018csrnet}, we use three dilated convolutional layers in the counting branch to enlarge the receptive fields. We incorporate feature maps from the localization branch with the counting branch to generate high-quality density maps. The combined features then go to three other convolutional layers. As a result, the spatial resolution of the last convolutional layer of the counting branch is 1/2 of the input image. Thus, we up-sample the output to the size of the input image using bi-linear interpolation to estimate the density map. The last convolutional layer of the counting branch has a linear activation function. To estimate the number of wheat heads at the inference time, we sum over the estimated density map.

\textbf{Localization branch:}

The localization branch estimates the localization map given an image of wheat heads where points on the localization map indicate the center points of wheat heads in an input image. The localization branch has four $1\times1$ convolutional layers to further extract relevant features for the localization map estimation. The spatial resolution of the last convolutional layer of the localization branch is 1/2 of the input image. As such, we up-sample the output of the last convolutional layer to the size of the input image using bi-linear interpolation to estimate the localization map. After up-sampling, we apply the sigmoid operation to estimate the localization map. We employ $1\times1$ convolution throughout our network to add more nonlinearity and make the proposed model lighter. To estimate the precise locations of wheat heads from the predicted localization maps during inference, we first apply an average pooling of size $3\times3$ to reduce the noises and increase the values of true peaks. Then, we select all local peaks as the predicted center of wheat heads.

\subsection{Network Loss}

We use two different loss functions to accomplish both the density-based counting and the localization tasks, which are as follows:\\

\textbf{Counting loss:} 

We use the Euclidean loss as defined in Equation \eqref{eq:eq_loss_density} for the density map estimation where the Euclidean loss computes the sum of the pixel-wise difference between the predicted and ground truth density maps for each image;

\begin{eqnarray}
L_{den}=\frac{1}{N}\sum_{i=1}^{N}\|\hat{D}(x_i)-D_i\|_{2}^{2}\label{eq:eq_loss_density}
\end{eqnarray}

\noindent where, $\hat{D}(x_i)$, $D_i$, $x_i$, and $N$ denote the predicted density map of the $i$th input image, the $i$th ground truth density map, the $i$th input image, and the number of images, respectively. Euclidean loss measures the distance between the estimated density map and the ground truth. 

\textbf{Localization loss:}

We employ pixel-wise Focal loss \citep{lin2017focal} as defined in Equation \eqref{eq:eq_loss_local} for the localization branch. Focal loss down-weights the loss assigned to well-classified examples which helps address the foreground-background class imbalance encountered during the training process.

\begin{eqnarray}
L_{loc}=-\frac{1}{N}\sum_{i=1}^{N}\left[\alpha\sum_{j=1}^{n\times m} \text{log}(\hat{\Phi}(x_i^j))\:(1-\hat{\Phi}(x_i^j))^\gamma \:\Phi(x_i^j)+ \text{log}(1-\hat{\Phi}(x_i^j))\:\hat{\Phi}(x_i^j)^\gamma \:(1-\Phi(x_i^j)) \right]\label{eq:eq_loss_local}
\end{eqnarray}

\noindent where, $\hat{\Phi}(x_i^j)$, $\Phi(x_i^j)$, $x_i$, $n$, $m$, and $N$ denote the predicted localization map of the $i$th input image at pixel $j$, the $i$th ground truth localization map at pixel $j$, the $i$th input image, the height of localization map, the width of localization map, and the number of images, respectively. $\gamma$ is the parameter of Focal loss which determines the rate at which easy examples are down-weighted. $\alpha$ is a balancing factor. We set  $\gamma = 2$ and $\alpha = 0.25$ in our experiments, respectively. Lastly, we define the final loss of the network as the weighted sum of counting and localization losses as $L=L_{den}+\beta \: L_{loc}$ with $\beta = 0.01$ selected after extensive numerical testing.

\section{Experiment and Results}\label{experiments}

This section presents the dataset used for our experiments, the evaluation metrics, the training hyper-parameters, and the final results. All experiments were conducted in Tensorflow \citep{abadi2016tensorflow} on an NVIDIA Tesla V100 GPU.

\subsection{Dataset}\label{sec:data_all}

The dataset being used is gathered from a publicly available wheat database from \cite{david2020global}. The 3,373 images were gathered in the RGB spectrum from 10 different locations around the world, including Europe, North America, Asia, and Australia. The row spacing of the wheat fields varies from 12.5 cm to 30.5 cm. In addition to variability in the row spacing, each field has different planting densities (number of seed drops per unit area). Moreover, the soil characteristics of the growth region vary from mountainous regions to traditional irrigated farmlands which leads to differences in color and lighting conditions. The images were captured using various cameras at different lengths from the wheat heads ranging from 1.8 m to 3 m. Given the drastically different growing environments and camera setups, each image is uniquely different as seen in Figure 1. For our study, we randomly selected 20\% of the dataset as test data and used the rest as the training data. Table \ref{tab:data_stat} shows the summary statistics of the training and test datasets.

 \begin{table}[H]
     \centering
     \begin{tabular}{|c|c|c|c|c|c|c|}
     \hline
      Dataset& Number of Images&  Resolution  & Min & Max& Avg & Total  \\
         \hline
      Training&    2,698& $1024\times 1024$  & 1 & 116 & 43.75 & 118,043\\ 
         \hline
           Test&    675& $1024\times1024$  & 1 & 97 & 43.96 & 29,679\\ 
           \hline
     \end{tabular}
     \caption{The statistics of dataset used in this study. Min, Max, Avg, and Total denote the minimum, maximum, average, and total number of annotated wheat heads, respectively. }
     \label{tab:data_stat}
 \end{table}


\subsubsection{Ground Truth Density Maps}

Let $P=\{p_1,..., p_M\}$ denote the annotation set of $M$ wheat heads where $p_i$ indicates the position of $i$th wheat head. Each wheat head can be represented by a delta function $\delta (x-p_i)$. Therefore, we can represent the ground truth for an image with $M$ wheat head annotations as follows:

\begin{eqnarray}
H(x)=\sum_{i=1}^{M}\delta(x-p_i)\label{eq:eq_1}
\end{eqnarray}

\noindent To generate the ground truth density map $D(x)$, $H(x)$ is convoluted with a Guassian kernel $G_\sigma(x)$ as follows: \\

\begin{eqnarray}
D(x)=\sum_{i=1}^{M}\delta(x-p_i) \ast G_\sigma(x)\label{eq:eq_density}
\end{eqnarray}

\noindent where $\sigma$ denotes the standard deviation, which is estimated based on the average distance of $k$-nearest neighboring head annotations. The ground truth density maps are generated with the property that summation over the density map is the same as the total number of wheat heads in the image. As a result, the proposed method learns to count the number of wheat heads as an auxiliary task while also learning how much each region of the image contributes to the total count \citep{khaki2021deepcorn}.

\subsubsection{Ground Truth Localization Maps}

Considering the annotation set $P$ of $M$ wheat heads as defined in Equations (3 \& 4), we can generate the ground truth localization maps as follows:

\begin{eqnarray}
\Phi(x)=\sum_{i=1}^{M}\delta(x-p_i) \ast K\label{eq:eq_local}
\end{eqnarray}

\noindent where $K=[0,1,0;1,1,1;0,1,0]$ is a $3\times3$ kernel which makes a very small neighborhood of each annotated wheat head to be classified as positive \citep{liu2019recurrent}.

\subsubsection{Data Augmentation}\label{sec:data}

We use the following data augmentation to generate sufficient data for training WheatNet. To make our proposed method robust against scale variations, we construct a multi-scale pyramidal representation \citep{boominathan2016crowdnet} considering scales of 0.4 to 1.0 incremented in steps of 0.2 times the original image resolution. Then, 9 patches of size $300\times300$ are randomly cropped from each scale of the image pyramid. We also used random flip and Gaussian noise augmentations on generated image patches.

\subsection{Evaluation Metrics}

To evaluate the performance of our proposed model, we use the mean absolute error (MAE) and root mean squared error (RMSE) metrics, which are defined as follows:\\

\begin{eqnarray}
MAE=\frac{1}{N}\sum_{i=1}^{N}|C_{i}^{pred}-C_{i}^{GT}|\label{eq:MAE_eq}
\end{eqnarray}

\begin{eqnarray}
RMSE=\sqrt{\frac{1}{N}\sum_{i=1}^{N}|C_{i}^{pred}-C_{i}^{GT}|^2}\label{eq:RMSE_eq}
\end{eqnarray}

\noindent where, $N$, $C_{i}^{pred}$, and $C_{i}^{GT}$ denote the number of test images, the predicted counting for $i$th image, and the ground truth counting for $i$th image, respectively.

\subsection{Training Details}

Our proposed method is trained end-to-end with the following parameters. Following data augmentation described in subsection \ref{sec:data}, we generated 97,128 image patches. We randomly select 90\% of generated patches as the training data and use the rest as the validation data to monitor the training process. The network weights are initialized with Xavier initialization \citep{glorot2010understanding}. We use stochastic gradient descent (SGD) with a mini-batch size of 16 to optimize the loss function using Adam optimizer \citep{kingma2014adam}. We do not use batch normalization \citep{ioffe2015batch} in our network. The initial learning rate is set to be $3\times10^{-4}$ which is gradually decayed to $1.5\times 10^{-6}$ during training. We train the proposed model for 130,000 iterations.

\subsection{Design of Experiments}

We compare WheatNet with seven state-of-the-art models to evaluate its counting performance. All the competing methods are applicable for object counting in an image, which are as follows:

\textbf{SCAR:} proposed by \cite{gao2019scar}, employs a CNN model with spatial/channel-wise attention modules for density-based counting. The spatial-wise attention module encodes the pixel-wise context of the entire image to improve the estimation of density maps at the pixel level. The channel-wise attention module extracts more discriminative features among different channels.

\textbf{ASD:} proposed by \cite{wu2019adaptive}, utilizes an adaptive scenario discovery framework to predict density map. Their proposed method uses two parallel sub-networks with different filter sizes to counter scale variations. Their proposed method also adaptively assigns weights to the output of two parallel sub-networks to find best the dynamic scenarios implicitly.

\textbf{SPN:} proposed by \cite{chen2019scale}, this model utilizes a CNN with a scale pyramid module to counter image scale variations. Their scale pyramid module uses different rates of dilated convolutions in parallel in high layers of the network.

\textbf{CSRNet:} proposed by \cite{li2018csrnet}, adopts a truncated VGG backbone for extraction of relevant features. Then, the extracted features go through a series of dilated convolutional layers as the back-end to enlarge receptive fields to replace pooling operations.

\textbf{SaCNN:} proposed by \cite{zhang2018crowd}, this model has a scale-adaptive network architecture that uses a VGG backbone for feature extraction. To counter scale variation, their proposed method combines feature maps from two different scales of the network.

\textbf{ Faster R-CNN:} proposed by \cite{ren2015faster}, this model is a region proposal based object detection method that uses a region proposal network that shares full-image
convolutional features with the detection network. We use Faster R-CNN with ResNet50 \citep{he2016deep} as the backbone which was pre-trained on COCO dataset \citep{lin2014microsoft}.

\textbf{ SSD:} proposed by \cite{liu2016ssd}, this model is a single-shot object detection method that uses a single deep neural network without requiring object
proposals. We use SSD with inception-V2 \citep{szegedy2016rethinking} which was pre-trained on COCO dataset \citep{lin2014microsoft}.

\subsection{Final Results}

After training our proposed model and all competing models on the training dataset, we evaluated their performance on the wheat head test dataset described in Section 3.1, which included 675 images of wheat heads. Table \ref{tab:result1} shows the counting performances of all models with respect to the evaluation metrics.

\begin{table}[H]
    \centering
    \begin{tabular}{|c|c|c|c|c|}
    \hline
        Method &  MAE & RMSE & Supervision&\begin{tabular}{c}
             Number of \\
             Parameters (M)\\
        \end{tabular}\\
         \hline
        
         SCAR \citep{gao2019scar}&5.93&7.45 &Point& 16.27\\
         \hline

         ASD \citep{wu2019adaptive}& 5.86&7.55&Point&50.86\\
         \hline
         SPN \citep{chen2019scale} & 5.91 &7.73&Point&32.41\\
             \hline
          CSRNet \citep{li2018csrnet}  & 6.61& 7.85&Point&16.26\\
        \hline

        SaCNN \citep{zhang2018crowd}  & 6.24 &7.95&Point&25.07 \\
        \hline
         Faster R-CNN \citep{ren2015faster}  & 4.25 &5.71&Box&43.17 \\

       \hline
         SSD \citep{liu2016ssd}  & 4.67 &5.97&Box&13.16 \\
         \hline
           Our proposed (localization branch)  & 4.28 & 5.77&Point&4.04 \\
       \hline
          Our proposed (counting branch)  & 3.91& 5.28&Point&4.04 \\
        \hline
          Our proposed (Avg)  & \textbf{3.85}& \textbf{5.19}&Point&\textbf{4.04} \\
        \hline
    \end{tabular}
    \caption{The performance comparison of the competing methods in wheat head counting on 675 test images.  In the proposed (Avg) model, the total number of estimated wheat heads is the average number of wheat heads computed by the counting branch and the localization branch.}\label{tab:result1}
\end{table}

\begin{figure}[H]
    \centering
    \includegraphics[scale=0.13]{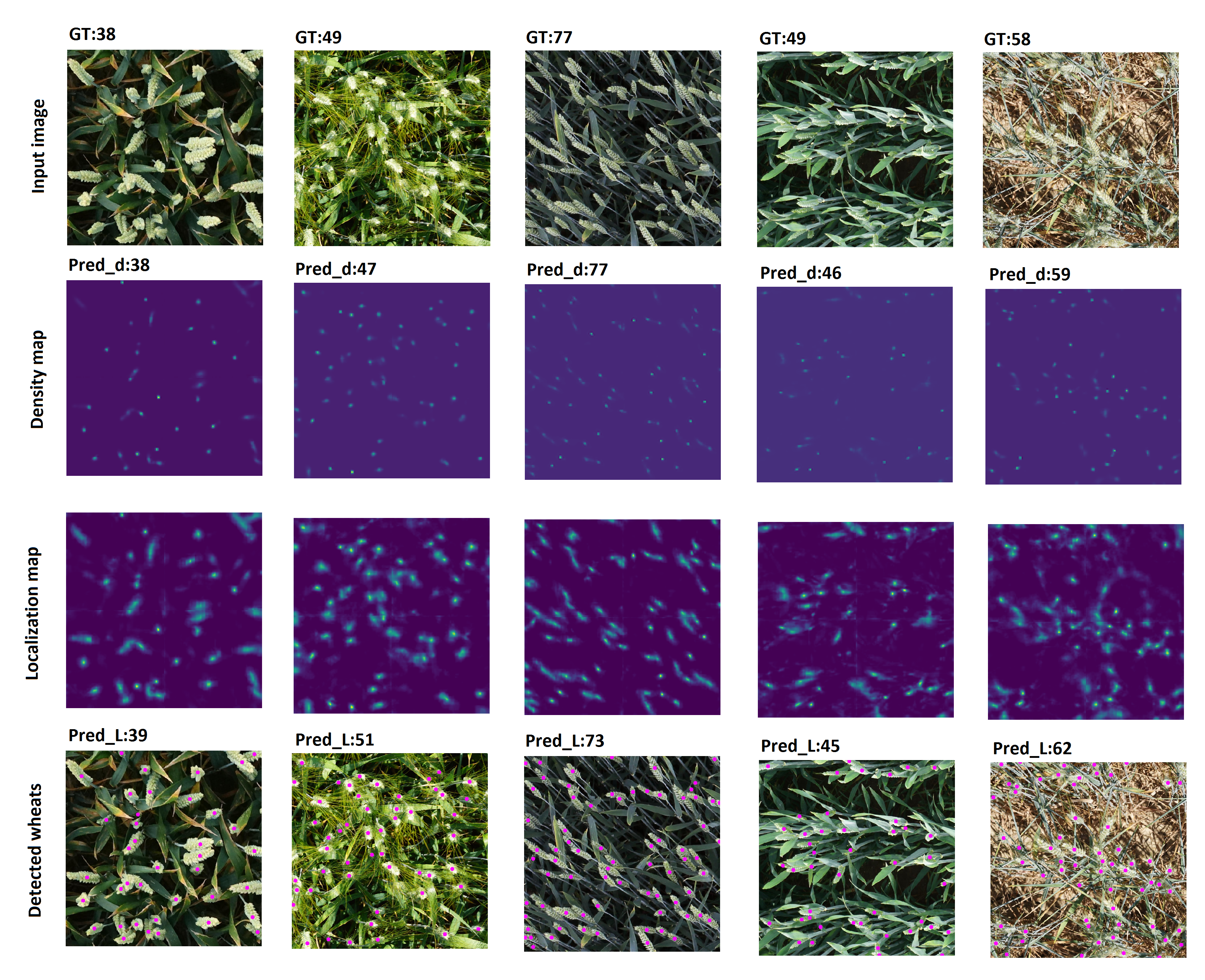}
    \caption{Visual results of our proposed method. The first, second, third, and fourth rows are, respectively, input images, estimated density maps, estimated localization maps, and detected wheat images. GT, $\text{Pred}_\text{d}$, and $\text{Pred}_\text{L}$ stand for the ground truth number of wheat heads, predicted number of wheat heads by counting branch, and predicted number of wheat heads by localization branch, respectively.}
    \label{fig:visual_res}
\end{figure}

As shown in Table \ref{tab:result1}, our proposed method outperformed other methods by varying extents. The counting branch had better counting performance compared to the localization branch because the counting branch is specifically designed to count the wheat heads based on an estimated density map. Among density based models, SCAR, ASD, and SPN had a comparable performance with SCAR having slightly lower prediction error than the other two. Object detection methods performed better than density based methods since some of the test images have sparse scenes of wheat heads and density-based methods are often well-suited for dense object counting. Faster R-CNN had a higher prediction accuracy than SSD due to using a region proposal network, however, SSD had a lower inference time compared to Faster R-CNN. The localization branch of our method outperforms all other methods except Faster R-CNN. Our proposed method had the highest prediction accuracy compared to other methods for the main following reasons: (1) the use of light backbone reduces the model's parameters, (2) the hybrid loss function of localization and density-based counting helps the model learn more discriminative features, (3) the Focal loss down-weights the loss assigned to well-classified examples which address the foreground-background class imbalance during the training process, and (4) the use of localization and density-based counting improves the performance of the model in both sparse and dense scenes. Figure \ref{fig:visual_res} visualizes a sample of the results of our proposed method that includes input images, predicted density maps, predicted localization maps, and detected wheat head images. Moreover, given our MAE of 3.85, we compute that our overall prediction error in terms of a percentage is $(3.85\times675)/29679 \approx 0.087$.

Our proposed method along with all density-based methods utilizes point-level annotations compared to object detection methods which use box-level annotations. The use of point-level annotation is beneficial due to being less labor-intensive, time consuming, and more practical for domains where publicly available annotated datasets are scarce. The inference time of our proposed method is 0.54 s on an Intel Core-i9 CPU 2.3 GHz. Compared to other methods, our proposed method has the lowest number of parameters and requires fewer computation resources.

\begin{figure}[H]
    \centering
    \includegraphics[scale=0.17]{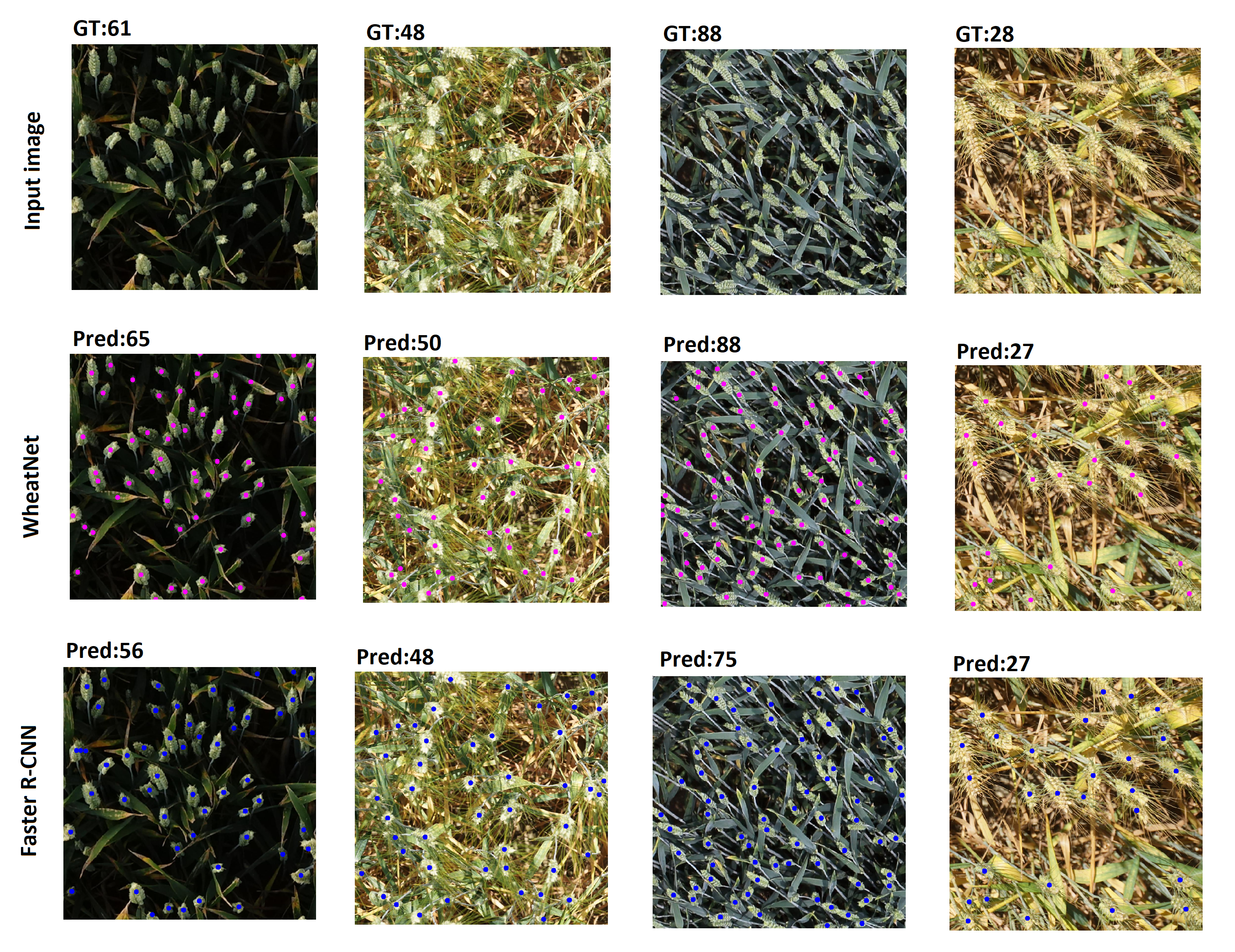}
    \caption{The predicted center of wheat heads estimated by our proposed method and Faster R-CNN method. The first, second, and third rows are, respectively, input images, detected wheat heads by WheatNet, and detected wheat heads by Faster R-CNN. GT and $\text{Pred}$ stand for the ground truth number of wheat heads and predicted number of wheat heads, respectively.}
    \label{fig:localization_res}
\end{figure}

To better evaluate the localization performance of our proposed method, we visualized the predicted center of wheat heads estimated by our proposed method and Faster R-CNN method. As shown in Figure \ref{fig:localization_res}, the localization performance of our proposed method is competitive with the Faster R-CNN method which is a state-of-the-art object detection method.
\newpage

\section{Analysis}

\subsection{Yield Estimation}

As previously stated, one of the key motivating factors of this work is the ability to perform live, in-field yield estimations before harvest begins. With this information, agronomists and farmers have the ability to make optimal grain management decisions (such as applying fertilization, irrigating, spraying for pests, etc.) to maximize their yield and, therefore, profitability. Traditionally, estimating wheat yield prior to harvest is done by manually counting the heads and then applying a well-established formula. It is commonly understood that there exists variability in a single estimate, so studies suggest doing replications through various points in the field. However, because of the sheer amount of wheat heads to count, manually identifying individual heads is time-consuming, labor-intensive, and prone to human error.

To aid in this approach, we couple our wheat head detection algorithm with a traditional approach provided by \cite{nebwheat}. The steps of this approach are as follows:

\begin{enumerate}
    \item Count the number of heads per foot of row and calculate the average number of heads per foot for the selected site.
    \item Count the number of kernels per head and calculate the average. Normally, there are 18-26 kernels per head, and assuming 22 is sufficient in practice.
    \item Measure the row distance in inches between the wheat varieties. This information is typically known by the farmer while planting seeds.
\end{enumerate}

\noindent In Equation 4, we provide the wheat yield estimation formula in bushels per acre as presented by \cite{nebwheat}.

\begin{equation}
    \textnormal{\textit{Wheat Yield}} = \dfrac{\dfrac{\#Heads}{Foot} \times \dfrac{\#Kernels}{Head}}{\textnormal{\textit{Row Spacing}}} \times 0.48
\end{equation}

\begin{figure}[H]
    \centering
    \includegraphics[scale=0.20]{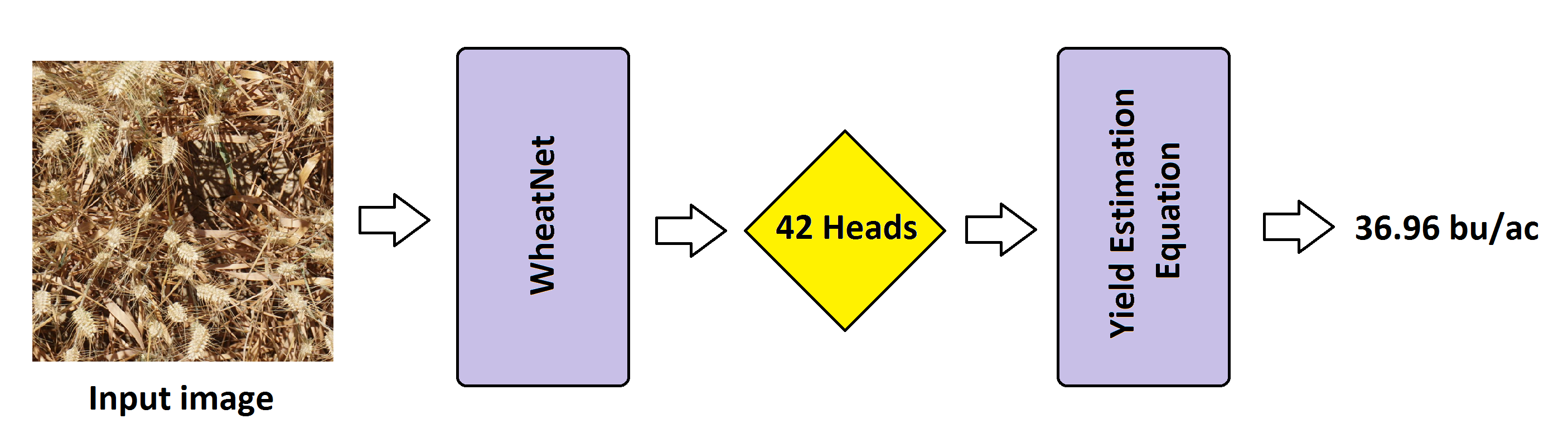}
    \caption{The wheat yield estimation procedure based on our proposed method. Estimated wheat heads and the amount of row spacing are 42 and 12 in, respectively.}
    \label{fig:yield_est}
\end{figure}

The multiplicative factor of 0.48 is a derived agronomic constant that incorporates kernel weight and unit conversion factors. In practice, small differences in the numbers used in the formula may result in large differences in yield estimates. Due to manual labor requirements, \cite{nebwheat} suggest repeating this process for five different field locations. However, with our proposed approach, easily hundreds of replications can be performed to get a more stable estimate of the true yield performance. Figure \ref{fig:yield_est} shows an example of wheat yield estimation based on our proposed method.

\section{Conclusion}

This paper presents a novel framework called WheatNet that allows for accurate and efficient counting of wheat heads from images. Our proposed network architecture makes use of a lightweight encoder and decoder which significantly reduces the network parameters. As a result, our proposed method can easily be deployed on mobile devices with limited computing power. We use a truncated MobileNetV2 as an encoder that merges feature maps to counter scale variations. Our proposed method employs two parallel branches for simultaneous density-based counting and localization using only point-level annotation. Our extensive quantitative results illustrate that WheatNet can indeed successfully localize and count the number of wheat heads regardless of lighting conditions and growing environment.  Coupled with our proposed wheat counting framework, we also illustrated a framework that can be used to estimate wheat yield using traditional agronomic formulas.

Given the fast inference time and size of our model, we are able to deploy our model for live, in-field usage on a mobile device to be used by farmers to aid in determining management practices to optimize crop growth. Historically, to acquire such data a farmer must traverse their field and manually count the heads in a given area. This approach is both time-consuming and error prone and must be replicated many times to get a consistent yield estimate. With WheatNet, farmers are now able to easily capture multiple, representative, field images to obtain an accurate and consistent yield estimate. We hope that our work helps pave the way for future work at the intersection of machine learning and agriculture to ultimately benefit plant science.

\section*{Conflicts of Interest}

The authors declare no conflict of interest.

\section*{Acknowledgement}

This work was partially supported by the National Science Foundation under the LEAP HI and GOALI programs (grant number 1830478) and under the EAGER program (grant number 1842097).

\bibliographystyle{unsrt}  
\bibliography{references}  


\end{document}